\documentclass[final,numberedheadings]{aipproc}
\layoutstyle{6x9}
\usepackage{amsmath, amssymb}
\begin{document}

	\title{Maximum Joint Entropy and Information-Based Collaboration of Automated Learning Machines}

	\classification{02.50Tt, 07.05.Bx, 07.07.Tw, 07.05.Fb, 07.05.Hd, 07.05.Kf.}
	\keywords{Joint Entropy, Collaboration, Information, Question, Automation.}

	\author{N. K. Malakar}{
		address={Department of Physics, University of Texas at Dallas}
	}

	\author{K. H. Knuth}{
		address={Department of Physics, University at Albany (SUNY)}
		,altaddress = {Department of Informatics, University at Albany (SUNY)}
	}
	\author{D. J. Lary}{
		address={Department of Physics, University of Texas at Dallas}
	}

\begin{abstract}
We are working to develop automated intelligent agents, which can act and react as learning machines with minimal human intervention.  To accomplish this, an intelligent agent is viewed as a question-asking machine, which is designed by coupling the processes of inference and inquiry to form a model-based learning unit.  In order to select maximally-informative queries, the intelligent agent needs to be able to compute the relevance of a question.  This is accomplished by employing the inquiry calculus, which is dual to the probability calculus, and extends information theory by explicitly requiring context.  Here, we consider the interaction between two question-asking intelligent agents, and note that there is a potential information redundancy with respect to the two questions that the agents may choose to pose.  We show that the information redundancy is minimized by maximizing the joint entropy of the questions, which simultaneously maximizes the relevance of each question while minimizing the mutual information between them.  Maximum joint entropy is therefore an important principle of information-based collaboration, which enables intelligent agents to efficiently learn together.	
	\end{abstract}
	\maketitle

\section*{INTRODUCTION}
Present day scientific explorations involve gathering data at an ever-increasing rate, thereby requiring autonomy as a vital part of exploration.   For example, remote science operations require automated systems that can both act and react with minimal human intervention.   Our vision is to construct an autonomous intelligent instrument system (AIIS) that collects data in an automated fashion, learns from that data, and then, based on the learning goal, decides which new measurements to take.  Such a system would constitute a learning machine that could act and react with minimal human intervention.  This is made possible by the comprehensive successes of Bayesian inference, the decision theoretic approach to experimental design \cite{fedorov1972theory,lindley_measure_1956, bernardo_expected_1979, Loredo03bayesianadaptive}, and
the development of the inquiry calculus \cite{cox_inference_1979,fry_cybernetics, knuth_what_2003, knuth_valuations_2006}.  

Our efforts to construct such autonomous systems \cite{knuth_arm_2007,knuth_center2010, Malakar2011} have considered the process of data collection and the process of learning in two distinct phases: the inquiry phase and inference phase.  By coupling these processes of inference and inquiry one can form a model-based learning unit that cyclically collects data and learns from that data by updating its models.  At this stage, the inference phase, which is based on Bayesian probability theory, is sufficiently well-understood so that our current focus is on inquiry.  For this reason, we tend to view the AIIS as a question-asking machine.

In this paper, we build upon our previous work \cite{knuth_arm_2007,knuth_center2010} and consider the problem of coordinating two question-asking intelligent agents.  Without coordination, after each agent has independently solved the presented problem, there will be a redundancy in the information obtained by the two agents.  In addition, during the question-asking process, there is a great potential for redundancy in terms of the questions that they pose.  We show that collectively this redundancy is minimized at each step of the question-asking process by maximizing the joint entropy of the two questions that the agents plan to ask.  This has the tendency to simultaneously maximize the relevance of each of the two questions posed while minimizing the mutual information between them.  We illustrate the process via simulation and show that maximization of the joint entropy is an important principle of information-based collaboration, which enables intelligent agents to efficiently learn together.



\section*{THE INQUIRY CALCULUS}
In this section, we briefly review the inquiry calculus.  The development of this calculus relies on several order-theoretic notions, which are more thoroughly discussed in papers outlining the theoretical development \cite{knuth_what_2003, knuth_valuations_2006}.  Central to this development is the concept of a partially ordered set, which is a set of elements in conjunction with a binary ordering relation.  Related, is a special case of a partially ordered set called a lattice, which is endowed with a pair of operations called the join and the meet so that the lattice can be thought of as an algebra where the join and meet are algebraic operators. Here we will consider elements that can be described in terms of sets.  So that the main concepts can be described in terms of subsets ordered by subset inclusion, the set union (join) and the set intersection (meet).

We consider three spaces: the state space, the hypothesis space, and the inquiry space.

The state space describes the possible states of the system itself. In the situations we will consider, the elements of the state space are mutually exclusive so that in terms of a partially ordered set, they can be represented as an antichain.

The hypothesis space describes what can be known about a system.  Its elements are sets of potential states of the system.  As such, it is a \emph{Boolean lattice} (or a Boolean algebra) constructed by taking the power set of the set of states and ordering them according to set inclusion.  In this space, the logical OR operation is implemented by set union (join) and the logical AND operation by set intersection (meet).  Logical deduction is straightforward in this framework, since implication is implemented by subset inclusion so that a statement in the lattice implies every statement that includes it in terms of subset inclusion.  General logical induction is implemented by quantifying degree to which one statement implies another with a real-valued bi-valuation, probability, which quantifies the degree to which one statement implies another.

The inquiry space describes what can be asked about a system.  Its elements are sets of statements, which are called questions, such that if a set contains a given statement, then it also contains all the statements that imply it.  In this sense, a question can be thought of as a set of potential statements that can be made.  It is constructed by taking down-sets of statements and ordering them by set inclusion resulting in a \emph{free distributive lattice}.  Just as some statements imply other statements, some questions answer other questions.  Specifically, if question $A$ is a subset of question $B$,  $A \subset  B$, then by answering question $A$, we will have necessarily answered question $B$.

Questions, which include all atomic statements as potential answers, are assured to be answerable by a true statement. Cox had termed such questions as real questions \cite{cox_inference_1979}.  If one considers the sub-space formed from the real questions, the minimal real question is defined as the central issue,
\begin{equation}
I = \bigcup_{i=1}^{n} X_i,
\end{equation}
where $X_i = \{\{x_i\}\}$ and $x_i$ is the statement `\textit{The system is in state $x_i$}'.  The central issue can then be expressed as the question `\textit{Is the system in state $x_1$ or in state $x_2$ ... or in state $x_n$?}'  Since it is the minimal real question, answering the central issue will necessarily answer all other real questions.

In practice, however, we cannot always pose the central issue directly.  A special class of real questions are the \emph{partition questions}, which partition the set of answers.  For example, given a set of atomic statements indexed by integers $1$ through $n$, we can partition this set in $p(n)$ ways, where $p(n)$ can be defined in terms of a generating function
\begin{equation}
\sum_{n=0}^{\infty}{p(n)x^n} = \prod_{k=1}^{\infty}\Big( \frac{1}{1-x^k} \Big),
\end{equation}
which blows up rapidly.
For example, for three atomic statements we have the set $\{1, 2, 3\}$ that can be partitioned as $\{1\}\{2\}\{3\}$ which results in the central issue
\begin{equation}
I = X_1 \cup X_2 \cup X_3.
\end{equation}
Another possible partitioning is $\{1\}\{2,3\}$, which represents the binary question `\textit{Is the system in state $x_1$ or not $x_1$?}' denoted
\begin{equation} \label{Eqn:P1|23}
P_{1|23} = X_1 \cup X_2X_3,
\end{equation}
where $X_1 = \{x_1\}$ and $X_2X_3 = \{x_2, x_3, x_2 \vee x_3\}$. In this way by answering $x_2$, $x_3$, or $x_2 \vee x_3 = \neg x_1$ one has provided the information that the system is not in state $x_1$.  Other partition questions are written similarly.

Valuations are handled in a way that is analogous to probability in the lattice of statements comprising the hypothesis space.  However, due to multiple competing constraints, a bi-valuation can only be consistently assigned to the partition sublattice of the real questions.  This bi-valuation is called the relevance, and is denoted $d(Q|P)$, which is read as `the degree to which $P$ answers $Q$' \cite{knuth_valuations_2006}.  In the special case where $P \subseteq Q$, we have that the relevance is maximal, which enables one to choose a grade so that $d(Q|P) = 1$.  Otherwise, the relevance takes on a value between $0$ and $1$.



The relevance of a partition question depends on the probability of its particular partition of answers.  One can show that the relevance of the question $P$ with respect to the central issue is given by the entropy of that partition of probabilities \cite{knuth_valuations_2006}.  In the case of the partition question $P_{1|23}$ described in (\ref{Eqn:P1|23}) we have that
\begin{equation}
d(I|P_{1|23}) \propto H(p_1, p_2+p_3)
\end{equation}
where $H$ is the Shannon entropy, and $p_i = Pr(x_i|\top)$ which is the probability that the system is in state $x_i$. 
The proportionality constant is the inverse of the relevance
\begin{equation}
d(I|I) \propto H(p_1, p_2, p_3)
\end{equation}
so that 
\begin{equation}
d(I|P_{1|23}) = \frac{H(p_1, p_2+p_3)}{H(p_1, p_2, p_3)}
\end{equation}
and
\begin{equation}
d(I|I) = \frac{H(p_1, p_2, p_3)}{H(p_1, p_2, p_3)} = 1.
\end{equation}



\section*{AUTOMATED EXPERIMENTAL DESIGN}
Previously, we  demonstrated a robotic arm, built with the LEGO MINDSTORMS NXT system, capable of autonomously locating and characterizing a white circle on a dark background \cite{knuth_arm_2007,knuth_center2010}.  Here we aim to extend this problem by introducing two robots that work in a collaborative effort to solve the same problem.

\subsection*{Computing with Questions}
The white circle is characterized by three unknown parameters $\{x_o, y_o, r_o\}$.  We are interested in asking questions about the center position of the circle $\{x_o, y_o\}$ as well as its radius $r_o$ by taking light intensity measurements centered at locations determined by the inquiry system.  Model-based descriptions enable one to make predictions about the outcomes of potential experiments. Given the joint posterior probability of the circle location and radius, one can determine the probability that a given intensity measurement at a position $(x_i,  y_i)$, will result in a ``white" or ``black" intensity reading.  This is easily done with sampling by maintaining a set of sampled circles and noting how many circles contain the proposed measurement location and would result in a white intensity reading, and how many circles do not contain the measurement circle resulting in a black intensity reading.  Such predictions can be made more precise by modeling the spatial sensitivity of the light sensor and computing the predicted numerical result of the sensor given the measurement location and the hypothesized characteristics of the white circle \cite{malakar_SSF2009}.  

Furthermore, the entropy associated with such a measurement can be computed as the entropy of the probability distribution of predicted measurement intensities.  This can be rapidly computed by generating a set of predicted measurements from the set of circles sampled from the posterior.  By generating a histogram of this set of predicted intensities, one has a model of the density function of predicted measurements.  The entropy of this histogram is computed and serves as an excellent estimate of the entropy associated with the question posed by recording the intensity at a particular measurement location.  By computing the entropy associated with a large set of measurement locations, one can create an entropy map based on the sampled circles and the known characteristics of the light sensor.  For increased speed, we also have developed an entropy-based search algorithm to intelligently search the entropy space without computing it everywhere \cite{Malakar2010}.

%



We begin by encoding the questions one might ask in terms of sets of circle parameters.  The central issue considers all possible circle parameter values, and in doing so asks the question ``Precisely where is the circle?''  In practice, this is a finite set since one can only measure to finite precision, and in the simulations we force it to be finite by considering a discrete grid of possible circle center positions and radii.  The central issue $I$ can be written as
\begin{equation}\label{circleQ}
  I = \left\{ \left\{ x_1,y_1,r_1\right\} ,\left\{ x_2,y_2,r_2\right\} ,...\right\},
\end{equation}
where each element of the set, such as $\{x_i, y_i, r_i\}$, represents a potential precise answer to the question.  One way to solve this problem is to simply ask all of the binary questions `Is the circle in state $\{x_i, y_i, r_i\}$?'  However, this is not very efficient.  Moreover, faced with measurement uncertainties, we do not know the exact answer to Eq. (\ref{circleQ}), as we cannot measure the exact values of the parameters of interest.  

Since we cannot directly perform a single, or even a small number, of measurements that directly answer the central issue.  Instead, we must identify measurements that can be performed that are maximally relevant to the central issue.  This involves finding measurement locations that have the maximum entropy as computed from the posterior probability of the circle states.





We note that any given measurement location $(x_{e1}, y_{e1})$ divides the space of circles into two regions: the set of circles that contain the measurement location, and the set of circles that do not contain the measurement location \begin{equation}
{Q}(x,y,r) = \left\{
 \bigcup_{
 \substack{
   (x_{e1},y_{e1}) \in circle
  }}\left\{ x_i,y_i,r_i\right\} ,
  \bigcup_{
 \substack{
   (x_{e1},y_{e1})\notin circle
  }}\left\{ x_j,y_j,r_j\right\}
\right\}.
\end{equation}
Similarly, a second robot choosing a different measurement location, $(x_{e2}, y_{e2})$, partitions the question space differently into two sets defining a different binary partition.  

Jointly, the two distinct measurement locations partition the space of circles into four regions, say $a$, $b$, $c$ and $d$: \begin{equation}
  a = \{white,~white\}, ~b = \{white,~black\}, ~c = \{black,~white\}, ~d=\{black,~black\} ,
\end{equation}
where, for example, $\{white, black\}$ refers to the set of circles that contain $(x_{e1},y_{e1})$ so that a measurement there will be predicted to result in a white intensity, but do not contain $(x_{e2},y_{e2})$ so that a measurement there will be predicted to result in a black intensity. The circles where the first robot measures white belong to the set $a \cup b$.  We can then define the elementary questions as \begin{align}
AB &= \{ a\lor b, a ,b\},
& CD = \{ c\lor d, c, d\},\nonumber \\
AC &= \{ a \lor c, a, c\},
& BD = \{ b\lor d, b, d \}.
\end{align}
and write the question that the first robot poses as \begin{equation}
 AB \cup CD = \{ a\lor b, c\lor d, a, b, c, d\} \equiv \{ \{w, \bullet \}, \{b,  \bullet \} \} ,
\end{equation}
and the question the second robot poses as
\begin{equation}
 AC \cup BD = \{ a\lor c, b\lor d, a, b, c, d\} \equiv \{ \{\bullet, w\}, \{ \bullet, b \} \} ,
\end{equation}
where the expressions on the right illustrate what the robots are measuring with $\bullet$ signifying either black or white.

Jointly the robots partition the space into four sets,\begin{equation}
  (AB \cup CD) \cap (AC \cup BD) =  (A \cup B \cup C \cup D).
\end{equation}
Therefore, the relevance of the joint question, with respect to the central issue, is given by the joint entropy of the predictions of the two measurements $E1 = (AB \cup CD)$ and $E2 = (AC \cup BD)$ \cite{Malakar2011} \begin{equation}
  d(I| E1 \cap E2)   = d(I | A \lor B \lor C \lor D)
    = H ( Pr(A), Pr(B), Pr(C), Pr(D) ),
\end{equation}
where, given that robot 1 measures at $(x_{e1}, y_{e1})$ and robot 2 measures at $(x_{e2},y_{e2})$,  $Pr(A)$ denotes the probability that both the first and the second robot's measurement locations result in a white intensity, $Pr(B)$ denotes the probability that the first measurement results in white and the second in black,  $Pr (C)$ denotes the probability that the first measurement results in black and the second in white, and  $Pr(D)$ denotes the probability that both the first and the second measurements result in black. Considered jointly, the predicted measurement results associated with the pair of measurement locations constitute a two-dimensional distribution at each point in the four-dimensional space of pairs of measurement locations.  The relevance $d(I|E1\cap E2)$ dictates that we select measurement locations that maximize the joint entropy of the intensities predicted to be measured by the two robots.


%


\section*{RESULTS}
In the present case of model based exploration, given a hypothesized circle location and radius, the intensity to be measured at any point in the field can be predicted. By considering 45 posterior samples, we made predictions about the intensities which gave a distribution of 45 predicted intensities. The entropy associated each possible measurement location was computed by estimating the entropy of the histogram of predicted intensities at that position in the field. This enables us to produce an entropy map for a single proposed measurement. Joint entropy maps would require four dimensions to display.  Instead, we plot the joint entropy of the two measurements for the case where the first experiment E1 is determined.  This map then represents a two-dimensional slice through the four-dimensional space of pairs of measurements.  The mutual information maps (not shown) can be made similarly.

\begin{figure}[h]
\centering
\includegraphics[width=1\textwidth]{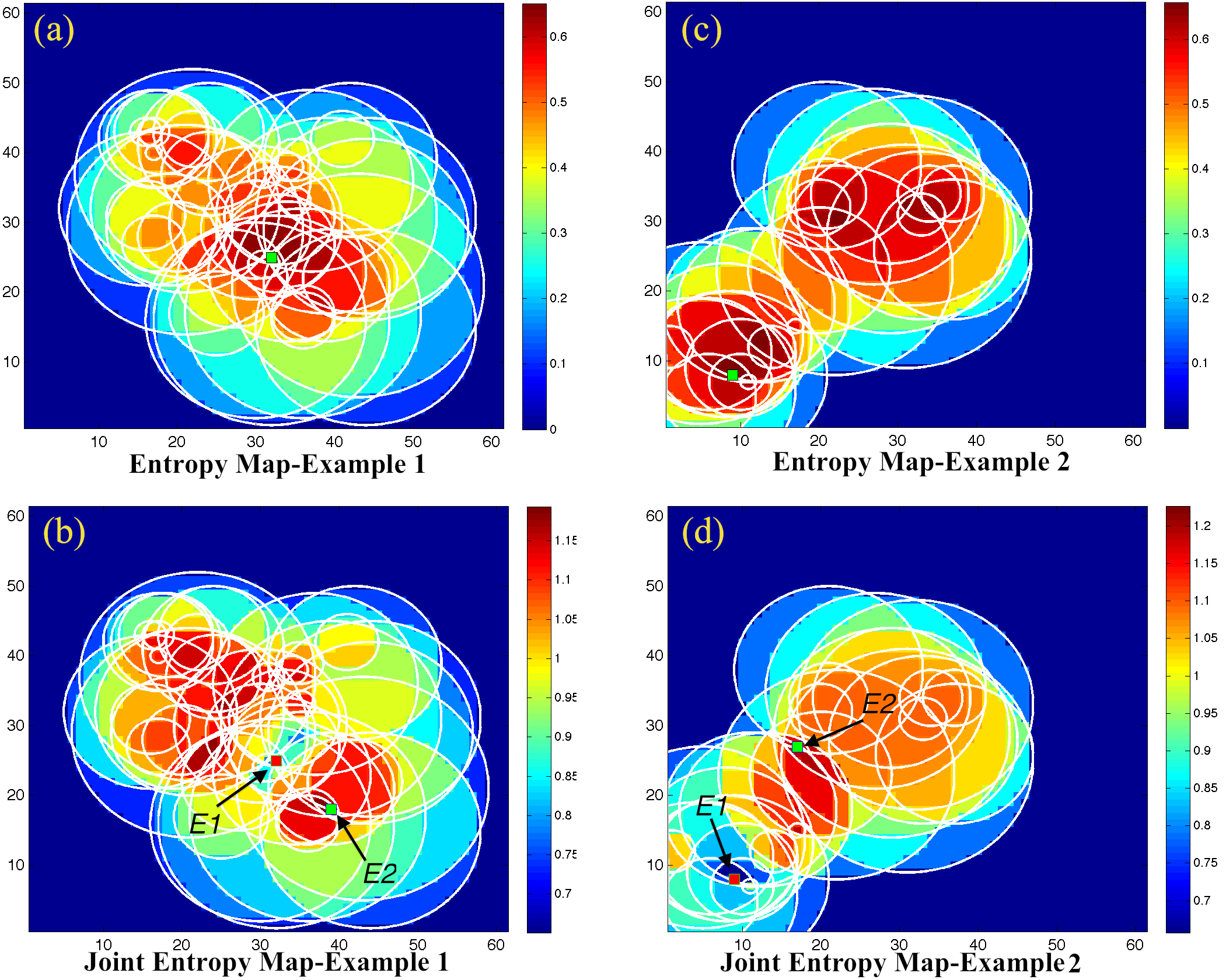}
\caption{Figures illustrating two set of examples where we implement the information-based collaboration for experimental design in the problem where two robots are to characterize a circle using light sensors. Figures (a) and (b) show the cases where circles are highly  correlated, whereas figures (c) and (d) show the cases where circles are less correlated. In both cases, we have drawn a set circles from the posterior samples and used these circles to make predictions about the expected measured light intensity at each point. The top figures (a and c) show the entropy map, which illustrates the optimal measurement location in the case where only one measurement is to be taken.  The botom figures (b and d) illustrate the joint entropy map of measurement location $E2$ shares with measurement location $E1$ fixed. Note that the selected location of meaurement $E2$ maximizes the joint entropy, which  involves finding an informative measurement location that does not provide information redundant to $E1$.}
\label{Figureall}
\end{figure}

Figure 1 shows two examples, where we considered different degrees of overlap of the sample circles. The figures on the left column represent the case with more correlated circles than those on the right. Each of the figures show a set of  circles drawn from the posterior probability. Overlaid on this are the  entropy maps (a and c),  and  the joint entropy maps  (b and d). 

The entropy maps in (Figures 1a and 1c) show the measurement location that would be selected in the event that only one measurement was being performed.  The joint entropy maps (Figure 1b and 1d) show the locations of the second measurement $E2$ that maximize the relevance of the question $d(I|E1 \cap E2)$ given that the location of $E1$ has been selected.  By comparing the locations of $E2$ in the joint entropy map with the corresponding values of entropy and mutual information, one can see that the selected measurement locations for $E2$ favor regions of high entropy while avoiding locations that share mutual information with $E1$.  Maximizing the joint relevance naturally chooses informative measurement locations that promise to provide independent information. The two measurement locations $E1$ and $E2$ that maximize the relevance of the joint question $d(I|E1 \cap E2)$ are indicated by arrows.



 \section*{CONCLUSIONS AND FUTURE APPLICATIONS}
In this paper, we have presented the method of information-based collaboration for Automated Intelligent Instruments System (AIIS). 
We have considered the intelligent agent as a question-asking machine and have focused on the inquiry phase, where our aim has been to select maximally informative queries with respect to a given goal. We have extended the order-theoretic approach \cite{knuth_what_2003, knuth_valuations_2006} to assign the relevance of questions for collaborative AIIS. We have shown that the joint entropy gives the relevance of the joint question posed by the agents. Maximum joint entropy is  an important principle of information-based collaboration, which enables intelligent agents to efficiently learn together.

Currently our team in UTD is working on to develop a fleet of aircrafts to deploy in the field using the technique of collaboration developed in this paper. The aircraft fleet consists of helicopters as well as the fixed wing small aircrafts.  We aim to use the fleet to characterize and help predict tornado forecasts, assist with the gas leak detection, and monitor the health of cattle. The work is in progress.

 \bibliographystyle{aipproc}
 \bibliography{nkm-me11}

\end{document}